\documentclass[11pt,a4paper]{article}
\usepackage[hyperref]{emnlp-ijcnlp-2019}
\usepackage{times}
\usepackage{latexsym}
\usepackage{pifont}
\usepackage{CJKutf8}
\usepackage[ruled,vlined]{algorithm2e}
\usepackage{smile}
\usepackage{amsmath}

\usepackage{setspace}
\AtBeginDocument{%
  \addtolength\abovedisplayskip{-0.5\baselineskip}%
  \addtolength\belowdisplayskip{-0.4\baselineskip}%
  \addtolength\abovedisplayshortskip{-0.5\baselineskip}%
  \addtolength\belowdisplayshortskip{-0.4\baselineskip}%
}

\usepackage{url}

\aclfinalcopy 

\newcommand{\RT}[1]{{\color{red}{#1}}}
\newcommand{\GT}[1]{{\color{blue}{#1}}}

\title{Contextual Text Denoising with Masked Language Models }

\author{Yifu Sun \thanks{ \quad Work done at Georgia Tech.} \\
  Tencent  \\
  {\tt yifusun2016@outlook.com} \\\And
  Haoming Jiang \\
  Georgia Tech  \\
  {\tt jianghm@gatech.edu} \\}

\date{}

\begin{document}
\maketitle
\begin{abstract}
\vspace{-0.05in}
Recently, with the help of deep learning models, significant advances have been made in different Natural Language Processing (NLP) tasks. Unfortunately, state-of-the-art models are vulnerable to noisy texts. We propose a new contextual text denoising algorithm based on the ready-to-use masked language model. The proposed algorithm does not require retraining of the model and can be integrated into any NLP system without additional training on paired cleaning training data. We evaluate our method under synthetic noise and natural noise and show that the proposed algorithm can use context information to correct noise text and improve the performance of noisy inputs in several downstream tasks.
\vspace{-0.05in}
\end{abstract}

\vspace{-0.1in}
\section{Introduction}
\vspace{-0.06in}

Based on our prior knowledge and contextual information in sentences, humans can understand noisy texts like misspelled words without difficulty. However, NLP systems break down for noisy text. For example, \citet{belinkov2017synthetic} showed that modern neural machine translation (NMT) system could not even translate texts with moderate noise. An illustrative example of English-to-Chinese translation using Google Translate \footnote{\url{https://translate.google.com}; Access Date: 08/09/2019} is presented in Table~\ref{tab:illustrate_example}.

\begin{table*}[!htb]
    \centering
    \begin{tabular}{c|c|c}
         \hline
         Method & Input Text & Google Translate  \\
         \hline
         \hline
         Clean Input & there is a fat \GT{duck} swimming in the \GT{lake} &  \begin{CJK}{UTF8}{gbsn}
        \GT{湖里} 有一只胖\GT{鸭子}在游泳
        \end{CJK} \\
         Noisy Input & there is a fat \RT{dack} swimming in the \RT{leake}  &  \begin{CJK}{UTF8}{gbsn}
        在 \RT{leake} 里游泳时有一个 \RT{胖子}
        \end{CJK} \\
         Spell-Checker  & there is a fat \RT{sack} swimming in the \RT{leak}  &  \begin{CJK}{UTF8}{gbsn}
        在 \RT{泄露处} 有一个肥胖 \RT{袋}在游泳
        \end{CJK} \\
        Grammaly\footnotemark & there is a fat \RT{dack} swimming in the \RT{lake} \textbf{} &  \begin{CJK}{UTF8}{gbsn}
         \RT{湖里} 游泳很胖
        \end{CJK} \\
         Ours & there is a fat \GT{duck} swimming in the \GT{lake} \textbf{} &  \begin{CJK}{UTF8}{gbsn}
         \GT{湖里} 有一只胖\GT{鸭子}在游泳
        \end{CJK} \\
         \hline
    \end{tabular}
    \vspace{-0.1in}
    \caption{Illustrative example of spell-checker and contextual denoising.
    \vspace{-0.2in}
}
    \label{tab:illustrate_example}
\end{table*}

Text correction systems are widely used in real-world scenarios to address noisy text inputs problem. Simple rule-based and frequency-based spell-checker are limited to complex language systems. More recently, modern neural Grammatical Error Correction (GEC) systems are developed with the help of deep learning  \citep{zhao2019improving,chollampatt2018neural}. These  GEC systems heavily rely on annotated GEC corpora, such as CoNLL-2014 \citep{ng2014conll}. The parallel GEC corpora, however, are expansive, limited, and even unavailable for many languages.  
Another line of researches focuses on training a robust model that inherently deals with noise.  For example, \citet{belinkov2017synthetic} train robust character-level NMT models using noisy training datasets, including both synthetic and natural noise. On the other hand, \citet{malykh2018robust} consider robust word vectors. These methods require retraining the model based on new word vectors or noise data. Retraining is expensive and will affect the performance of clean text. For example, in \citet{belinkov2017synthetic}, the robustness scarifies the performance of the clean text by about $7$ BLEU score on the EN-FR translation task.

In this paper, we propose a novel text denoising algorithm based on the ready-to-use masked language model (MLM, \citet{devlin2018bert}). Notice that we are using English Bert. For other languages, We need to use MLM model pre-trained on that specific language. The design follows the human cognitive process that humans can utilize the context, the spell of the wrong word \citep{mayall1997disruption},  and even the location of the letters on the keyboard to correct noisy text. The MLM essentially mimics the process that the model predicts the masked words based on their context. There are several benefits of the proposed method:
\begin{itemize}
    \vspace{-0.12in}
    \item Our method can make accurate corrections based on the context and semantic meaning of the whole sentence as Table~\ref{tab:illustrate_example}~shows.
    \vspace{-0.12in}
    \item The pre-trained masked language model is ready-to-use \citep{devlin2018bert,liu2019roberta}. No extra training or data is required. 
    \vspace{-0.12in}
    \item Our method makes use of Word Piece embeddings \cite{wu2016google} to alleviate the out-of-vocabulary problem.
    \vspace{-0.08in}
\end{itemize}

\vspace{-0.05in}
\section{Method}
\vspace{-0.05in}


Our denoising algorithm cleans the words in the sentence in sequential order. Given a word, the algorithm first generates a candidate list using the MLM and then further filter the list to select a candidate from the list. In this section, we first briefly introduce the masked language model, and then describe the proposed denoising algorithm. 

\vspace{-0.05in}
\subsection{Masked Language Model}

\footnotetext{\url{https://app.grammarly.com}; Access Date: 08/09/2019}
Masked language model (MLM) masks some words from a sentence and then predicts the masked words based on the contextual information. Specifically, given a sentence $\bx=\{x_i\}_{i=1}^L$ with $L$ words, a MLM models 
\begin{align*}
    p(x_j|x_1,...,x_{j-1},[MASK],x_{j+1},...,x_L),
\end{align*}
where $[MASK]$ is a masking token over the $j$-th word. Actually, MLM can recover multiple masks together, here we only present the case with one mask for notation simplicity. In this way, unlike traditional language model that is in left-to-right order (i.e., $p(x_j|x_1,...,x_{j-1})$), MLM is able to use both the left and right context. As a result, a more accurate prediction can be made by MLM. In the following, we use the pre-trained masked language model, BERT \cite{devlin2018bert}. So no training process is involved in developing our algorithm.

\vspace{-0.05in}
\subsection{Denoising Algorithm}


The algorithm cleans every word in the sentence with left-to-right order except for the punctuation and numbers by masking them in order. For each word, MLM first provide a candidate list using a transformed sentence. Then the cleaned word is selected from the list.
The whole process is summarized in Algorithm~\ref{alg:main}.

\noindent\textbf{Text Masking} The first step is to convert the sentence $\bx$ into a masked form $\bx'$. With the use of Word Piece tokens, each word can be represented by several different tokens. Suppose the $j$-th word (that needs to be cleaned) is represented by the $j_s$-th token to the $j_e$-th token, we need to mask them out together. For the same reason, the number of tokens of the expected cleaned word is unknown. So we use different number of masks to create the masked sentence $\{\bx_n'\}_{n=1}^N$, where $\bx_n'$ denotes the masked sentence with $n$-gram mask. Specifically, given $\bx=x_1,...,x_{j_s},...,x_{j_e},...,x_L$, the masked form is $\bx_n'= x_1,...,[MASK]\times n,...,x_L$. We mask each word in the noisy sentence by order. The number of masks $N$ can not be too small or too large. The candidate list will fail to capture the right answer with a small $N$. However, the optimal answer would fit the noisy text perfectly with a large enough $N$. Empirically, we find out $N=4$ is sufficiently large to obtain decent performance without too much overfitting.

\noindent\textbf{Text Augmentation} Since the wrong word is also informative, so we augment each masked text $\bx_n'$ by concatenating the original text $\bx$. Specifically, the augmented text is $\tilde{\bx}_n = \bx_n' [SEP] \bx$, where $[SEP]$ is a separation token.\footnote{In BERT convention, the input also needs to be embraced with a $[CLS]$ and a $[SEP]$ token.}

Compared with directly leaving the noisy word in the original sentence, the masking and augmentation strategy are more flexible. It is benefited from that the number of tokens of the expected word does not necessarily equal to the noisy word. Besides, the model pays less attention to the noisy words, which may induce bias to the prediction of the clean word.


\noindent\textbf{Candidate Selection}
The algorithm then constructs a candidate list using the MLM, which is semantically suitable for the masked position in the sentence. We first construct candidate list $V_c^n$ for each $\tilde{\bx}_n$, and then combine them to obtained the final candidate list $V_c = V_c^1 \cup \cdots \cup V_c^N$. Note that we need to handle multiple masks when $n>1$. So we first find $k$ most possible word pieces for each mask and then enumerate all possible combinations to construct the final candidate list.  Specifically, 
\begin{align*}
    V_c^n=\textrm{Top-}k\{p([MASK]_1=w|\tilde{\bx}_n)\}_{w \in V} \\
    \times \cdots \times \textrm{Top-}k\{p([MASK]_n=w|\tilde{\bx}_n)\}_{w \in V},
\end{align*}
where $V$ is the whole vocabulary and $\times$ means the Cartesian product. 

There may be multiple words that make sense for the replacement. In this case, the spelling of the wrong word is useful for finding the most likely correct word. We use the edit distance to select the most likely correct word further.
\begin{align*}
    w_c = \arg\min_{w\in V_c} E(w, x_j),
\end{align*} where $E(w, x_j)$ represent the edit distance between $w$ and the noisy word $x_j$.

\begin{algorithm}[h]
\SetAlgoLined
\KwIn{Noisy sentence $\bx=\{x_i\}_{i=1}^L$}
\KwOut{Denoised sentence $\bx=\{x_i\}_{i=1}^L$}
 \For{$i = 1,2,...,L$ }{
    $\{\bx_n'\}_{n=1}^N = \textrm{Masking}(\bx)$ \;
    $\{\tilde{\bx}_n\}_{n=1}^N = \{ \textrm{Augment}(\bx_n',\bx)    \}_{n=1}^N$ \;
    \For{$n = 1,2,...,N$}{
        $V_c^n = \textrm{Candidate}(\tilde{\bx}_n)$ \;
    }
    $V_c = V_c^1 \cup \cdots \cup V_c^N$ \;
   $w_c = \arg\min_{w\in V_c} E(w, x_j)$ \;
  $x_i = w_c$\;
 }
 \caption{Denoising with MLM }
 \label{alg:main}
\end{algorithm}

\vspace{-0.1in}
\section{Experiment}

We test the performance of the proposed text denoising method on three downstream tasks: neural machine translation, natural language inference, and paraphrase detection. All experiments are conducted with NVIDIA Tesla V100 GPUs. We use the pretrained pytorch Bert-large (with whole word masking) as the masked language model \footnote{\url{https://github.com/huggingface/pytorch-pretrained-BERT}}. For the denoising algorithm, we use at most $N=4$ masks for each word, and the detailed configuration of the size of the candidate list is shown in Table~\ref{tab:can_size}. We use a large candidate list for one word piece which covers the most cases. For multiple masks, a smaller list would be good enough.


For all tasks, we train the task-specific model on the original clean training set. Then we compare the model performance on the different test sets, including original test data, noise test data, and cleaned noise test data.  
We use a commercial-level spell-checker api \footnote{\url{https://rapidapi.com/montanaflynn/api/spellcheck}; Access Date: 08/09/2019} as our baseline method.

\begin{table}[]
    \centering
    \begin{tabular}{c|cc}
        \hline
       No. of $[MASK]$ $(n)$ & Top  $k$ & Size \\
       \hline
       \hline
        $1$ & $3000$ & $3000$ \\
        $2$ & $5$ & $25$ \\
        $3$ & $3$ & $27$ \\
        $4$ & $2$ & $16$ \\
        \hline 
        Total: & & $3068$ \\
        \hline 
    \end{tabular}
    \caption{Size of the candidate list}
    \label{tab:can_size}
    \vspace{-0.2in}
\end{table}

In this section, we first introduce how the noise is generated, and then present experimental results of three NLP tasks. 

\vspace{-0.05in}
\subsection{Noise}
To control the noise level, we randomly pick words from the testing data to be perturbed with a certain probability. For each selected word, we consider two perturbation setting: artificial noise and natural noise. 
Under \textit{artificial noise} setting, we separately apply four kinds of noise: {Swap}, {Delete}, {Replace}, {Insert} with certain probability. Specifically,
\begin{itemize}
    \vspace{-0.1in}
    \item Swap: We swap two letters per word. 
    \vspace{-0.1in}
    \item Delete: We randomly delete a letter in the middle of the word. 
    \vspace{-0.1in}
    \item Replace:  We randomly replace a letter in a word with another.
    \vspace{-0.1in}
    \item Insert: We randomly insert a letter in the middle of the word.
\end{itemize}

Following the setting in \cite{belinkov2017synthetic}, the first and the last character remains unchanged.

For the \textit{artificial noise}, we follow the experiment of \citet{belinkov2017synthetic} that harvest naturally occurring errors (typos, misspellings, etc.) from the edit histories of available corpora. It generates a lookup table of all possible errors for each word. We replace the selected words with the corresponding noise in the lookup table according to their settings.



\vspace{-0.05in}
\subsection{Neural Machine Translation}

We conduct the English-to-German translation experiments on the TED talks corpus from IWSLT 2014 dataset \footnote{\url{https://wit3.fbk.eu/archive/2014-01/texts/en/de/en-de.tgz}}. The data contains about $160,000$ sentence pairs for training, $6,750$ pairs for testing.

We first evaluate the performance using a 12-layer transformer implemented by fairseq \cite{ott2019fairseq}. For all implementation details, we follow the training recipe given by fairseq \footnote{\url{https://github.com/pytorch/fairseq/tree/master/examples/translation}}. We also evaluate the performance of Google Translate.

For the artificial noise setting, we perturb $20\%$ words and apply each noise with probability $25\%$. For that natural noise setting, we also perturb $20\%$ words. All experiment results is summarized in Table~\ref{tab:resultnmt}, where we use BLEU score \cite{papineni2002bleu} to evaluate the translation result.

\begin{table}[ ht]
\centering
    \begin{tabular}{c|c c}
         \hline
         Text Source & Google & Fairseq \\
         \hline
         \hline
         Original & $31.49$ & $28.06$\\
         \hline
         Artificial Noise & $28.11$ & $22.27$ \\
         + Spell-Checker & $26.28$ & $21.15$ \\
         + Ours & $\mathbf{28.96}$ & $\mathbf{25.80}$ \\
         \hline
         Natural Noise & $25.22$ & $17.29$ \\
         + Spell-Checker & $20.90$ & $15.04$ \\
         + Ours & $\mathbf{25.49}$ & $\mathbf{21.40}$ \\
        \hline 
    \end{tabular}
    \caption{BLEU scores of EN-to-DE tranlsation}
    \label{tab:resultnmt}
    \vspace{-0.1in}
\end{table}

As can be seen, both fairseq model and Google Translate suffer from a significant performance drop on the noisy texts with both natural and synthetic noise. When using the spell-checker, the performance even drops more.  Moreover, our purposed method can alleviate the performance drop. 







\vspace{-0.05in}
\subsection{Natural Language Inference}

We test the algorithm on Natural Language Inference (NLI) task, which is one of the most challenge tasks related to the semantics of sentences.
We establish our experiment based on the SNLI (the Stanford Natural Language Inference, \citet{bowman2015large}) corpus. Here we use accuracy as the evaluation metric for SNLI.


Here we use state-of-the-art 400 dimensional Hierarchical BiLSTM with Max Pooling (HBMP) \cite{talman2019sentence}. The implementation follows the publicly released code \footnote{\url{https://github.com/Helsinki-NLP/HBMP}}. We use the same noise setting as the NMT experiments. All results are presented in Table~\ref{tab:resultsnli}. We observe performance improvement with our method. To see if the denoising algorithm would induce noises to the clean texts, we also apply the algorithm to the original sentence and check if performance will degrade. It can be seen that, unlike the traditional robust model approach, applying a denoising algorithm on a clean sample has little influence on performance.

\begin{table}[!htb]
\centering
    \begin{tabular}{c|ccccc}
        \hline 
         \multirow{2}{*}{Method}  & \multirow{2}{*}{Original} & Artificial  & Natural   \\
          &  &  Noise &  Noise  \\
         \hline
        \hline 
         HBMP & $84.0$ & $75.0$ & $74.0$ \\
         +Spell-Checker & $84.0^*$ & $63.0$ & $68.0$ \\
         +Ours & $83.0^*$  &  $\mathbf{81.0}$ & $ \mathbf{77.0}$ \\ 
        \hline 
    \end{tabular}
    \caption{SNLI classification accuracy with artificial noise and natural noise. $^*$: Applying denoising algorithm on original texts. }
    \label{tab:resultsnli}
    \vspace{-0.2in}
\end{table}

As shown in the Table\ref{tab:resultsnli}, the accuracy is very close to the original one under the artificial noise. Natural noises contain punctuations and are more complicated than artificial ones. As a result,  inference becomes much harder in this way.




\vspace{-0.1in}
\subsection{Paraphrase Detection}
We conducted Paraphrase detection experiments on the Microsoft Research Paraphrase Corpus (MRPC, \citet{dolan2005automatically}) consisting of 5800 sentence pairs extracted from news sources on the web. It is manually labelled for presence/absence of semantic equivalence.

We evaluate the performance using the state-of-the-art model: fine-tuned RoBERTa \cite{liu2019roberta}. For all implemented details follows the publicly released code \footnote{\url{https://github.com/pytorch/fairseq/tree/master/examples/roberta}}. All experiment results is summarized in Table \ref{tab:resultmrpc}. We increase the size of the candidate list to $10000+25+27+16=10068$ because there are a lot of proper nouns, which are hard to predict.

\begin{table}[!htb]
\centering
    \begin{tabular}{c|ccccc }
        \hline 
         \multirow{2}{*}{Method}  & \multirow{2}{*}{Original} & Artificial  & Natural   \\
          &  &  Noise &  Noise  \\
        \hline 
         \hline
         RoBERTa & 84.3 & 81.9 & 75.2 \\
         +Spell-Checker & 82.6 & 81.3 & 75.4\\
         +Ours & {83.6}  &  \textbf{82.7} & \textbf{76.4} \\
        \hline 
    \end{tabular}
    \caption{Classification F1 score on MRPC }
    \label{tab:resultmrpc}
    \vspace{-0.1in}
\end{table}


\vspace{-0.2in}
\section{Conclusion and Future Work}
\vspace{-0.03in}
In this paper, we present a novel text denoising algorithm using ready-to-use masked language model. We show that the proposed method can recover the noisy text by the contextual information without any training or data. We further demonstrate the effectiveness of the proposed method on three downstream tasks, where the performance drop is alleviated by our method.  
A promising future research topic is how to design a better candidate selection rule rather than merely using the edit distance. We can also try to use GEC corpora, such as CoNLL-2014, to further fine-tune the denoising model in a supervised way to improve the performance.



\bibliographystyle{acl_natbib}
\bibliography{emnlp-ijcnlp-2019}

\end{document}